\documentclass[fleqn,10pt]{wlscirep}
\usepackage[utf8]{inputenc}
\usepackage[T1]{fontenc}
\usepackage{subfig}
\usepackage{booktabs}
\usepackage{multirow}
\usepackage{float}
\usepackage{graphicx}

\title{TRON: Transferring Naturally Evolved Network Architectures into Neuroevolution}

\author[2,*]{Zacharopoulos, Konstantinos}
\author[4]{Bekou, Alexandra}
\author[1]{Serradell, Jose M}
\author[4]{Paraskevopoulou Foteini}
\author[4]{Fournaris, Apostolos}
\author[3+]{Arapakis, Ioannis}
\author[1,*,+]{Lao, Oscar}

\affil[1]{IBE, Institute of Evolutionary Biology (CSIC-Universitat Pompeu Fabra), Barcelona, 08003, Spain}
\affil[2]{Technical University of Crete, ECE Department,Chania, 732 00, Greece}
\affil[3]{Barcelona Supercomputing Center, Barcelona, 08034, Spain}
\affil[4]{R.C. ATHENA, Industrial Systems Institute, Patras, 265 04, Greece}
\affil[5]{Computer Science Department, University of Turin, Turin, 10149, Italy}
\affil[6]{ISI Foundation, Turin, 10126, Italy}
\affil[*]{constantine.zach97@gmail.com, oscar.lao@ibe.upf-csic.es}
\affil[+]
{both authors equally contributed}


\begin{abstract}

Neural architecture search (NAS) remains a challenging optimization problem due to the vast and highly non-linear space of possible artificial neural network (ANN) topologies. Evolutionary approaches have demonstrated the ability to automatically generate ANN architectures; however, their effectiveness is often limited by the difficulty of preserving functional organization under mutation and recombination. In biological systems, complex networked phenotypes emerge from genetic information that has remained evolvable over long evolutionary timescales, suggesting that naturally evolved biological networks may encode organizational principles that facilitate adaptation.

Here, we present TRON (TRansfer Of Natural networks), a neuroevolution framework that incorporates experimentally characterized protein-protein interaction (PPI) networks as architectural priors for ANN evolution. Rather than initializing search from randomly generated topologies, TRON exploits biologically derived interaction patterns as reusable structural building blocks. The framework combines classical evolutionary operators with mechanisms inspired by neuroevolution, including speciation, while introducing a species quarantine strategy designed to protect newly generated architectures during an initial maturation phase. 

We evaluated TRON on three classification benchmarks (CORA, CITESEER, and PUBMED) and three regression benchmarks (DIABETES, FRIEDMAN, and CALIFORNIA), comparing its performance against established evolutionary and neuroevolutionary approaches. Across all classification datasets and most regression tasks, TRON achieved the highest predictive performance among the evaluated methods. Furthermore, the framework exhibited substantially lower variability across independent runs, indicating improved robustness and convergence stability during evolutionary search. While these gains were associated with increased computational requirements, the results consistently suggest that biologically derived interaction networks provide more favorable substrates for neuroevolution than randomly generated topologies. 

Taken together, our findings support the hypothesis that protein interaction networks can serve as effective architectural priors for ANN evolution. More broadly, this work demonstrates the feasibility of transferring naturally evolved biological network architectures into artificial intelligence systems and suggests a new direction for improving evolvability in neural architecture search through the reuse of biological organizational principles. \end{abstract}
\begin{document}

\flushbottom
\maketitle

\thispagestyle{empty}

\noindent Please note: Abbreviations should be introduced at the first mention in the main text --- no abbreviations lists. Suggested structure of main text (not enforced) is provided below.

\section*{Introduction}

Artificial neural networks (ANNs) are a family of machine learning models inspired by the organization of biological nervous systems. The architecture of an ANN, defined by the arrangement of its neurons and connections, plays a central role in determining its representational capacity, generalization performance, and computational efficiency \cite{LeCun2015}. Identifying effective architectures, however, remains a challenging optimization problem due to the enormous number of possible network topologies. 

To address this challenge, neural architecture search (NAS) methods have been developed to automate the design of ANN architectures \cite{elsken2019neural}. Among these approaches, evolutionary algorithms (EAs) provide a particularly attractive framework because they naturally support the exploration of large, discrete, and non-differentiable search spaces. In neuroevolution, evolutionary principles such as mutation, recombination, and selection are applied directly to neural networks, enabling the optimization of network parameters, architectures, or both simultaneously \cite{Kenneth2002, floreano2008neuroevolution}. Approaches such as NeuroEvolution of Augmenting Topologies (NEAT) demonstrated that evolutionary search can successfully discover increasingly complex neural architectures while maintaining architectural diversity through mechanisms such as speciation \cite{Kenneth2002}.

Despite these successes, neuroevolution still faces important challenges. Small syntactic modifications to an encoded neural network may produce disproportionately large functional changes in the resulting architecture. As a consequence, mutation and recombination frequently disrupt useful organizational structures, limiting the efficiency of evolutionary search. This problem becomes increasingly severe as network complexity increases and represents a longstanding challenge for evolutionary representations of neural systems. 

Biological evolution provides an intriguing source of inspiration for addressing this limitation. In living organisms, complex networked phenotypes emerge from genetic information that has remained evolvable over hundreds of millions of years of evolution. However, directly emulating the genetic encoding of biological nervous systems is difficult because the mapping between DNA sequence and mature neural circuitry involves multiple intermediate processes, including gene regulation, development, cellular differentiation, and environmental interactions. 

An alternative perspective is to focus on biological networks whose organization is more directly related to their genetic encoding. Protein-protein interaction (PPI) networks represent one such example. Proteins are the primary functional products of genes, and their interactions depend largely on sequence-derived structural features. Consequently, the relationship between genotype and network organization is substantially more direct than for large-scale nervous systems. 

Moreover, PPI networks frequently exhibit modular and hierarchical organization, properties that have long been proposed to contribute to robustness, functional specialization, and evolvability in biological systems \cite{hartwell1999molecular, barabasi2004network, wagner2007road}. Experimental and computational studies have shown that cellular functions are often organized into semi-independent modules whose components interact more strongly within modules than between them, facilitating the reuse and adaptation of functional units across different biological contexts \cite{hartwell1999molecular, spitkovsky2010modular}. 

Importantly, large collections of experimentally characterized PPI networks are now available across diverse organisms. Advances in high-throughput experimental techniques, together with extensive curation efforts, have led to the reconstruction of protein interaction maps containing thousands of proteins and interactions in organisms ranging from yeast to humans \cite{stelzl2005human, rolland2014proteome, luck2020reference}. These resources can be viewed as biological interaction architectures that have persisted under evolutionary pressures and therefore represent a rich source of naturally occurring connectivity patterns. While it remains unclear which specific properties contribute to their evolutionary persistence, the reuse of such structures may provide useful architectural priors for evolutionary search. Rather than initializing neuroevolution from random topologies, biologically derived interaction motifs may bias the search process toward more coherent regions of the architectural landscape and reduce the disruptive effects of mutation and recombination.

In this study, we propose TRON, a neuroevolution framework that incorporates experimentally characterized PPI networks as architectural substrates for artificial neural network evolution. Rather than searching exclusively from randomly generated topologies, TRON reuses biologically derived interaction patterns as candidate building blocks for ANN architectures. We hypothesize that these structures may provide a more evolvable substrate for neuroevolution by promoting the preservation of functional organization during mutation and recombination. To evaluate this hypothesis, we compare TRON against established evolutionary and neuroevolutionary approaches across a collection of classification and regression benchmarks and assess its impact on both predictive performance and evolutionary robustness.

\section{Related Work}
Gradient-based optimisation is highly effective for training the weights of a fixed network, but it is poorly suited to searching over architectures themselves, leaving  design decisions such as depth, width and connectivity to be chosen manually. Differentiable NAS methods such as DARTS \cite{Liu2019} avoid this by relaxing the architecture choice into a continuous mixture of candidate operations. This makes the search efficient, but it confines it to predefined cell-based search spaces and makes it prone to collapse towards degenerate architectures \cite{zela2020understanding}. However, the resulting topologies are stacks of repeated, hand-designed cells. They carry no information about which structures stay functional when modified, a property that biological networks have acquired under long-term evolutionary pressure. Where biological networks have entered deep learning, they appear either as input data, as in graph convolutional networks \cite{kipf2017semi}, whose architectures remain hand-designed and suffer from over-smoothing \cite{li2018deeper}, or as fixed wiring, as in knowledge-primed models such as P-NET \cite{elmarakeby2021biologically} and KPNNs \cite{fortelny2020knowledge}, whose topologies are imposed rather than searched. Neither treats biological interaction patterns as reusable, evolvable building blocks.

Population-based metaheuristics offer a natural alternative. They require no gradients, and because they maintain a population of candidates rather than a single solution, they can explore several regions of the search space in parallel. Among these optimisers, Particle Swarm Optimization (PSO) \cite{kennedy1995particle} is one of the earliest and most widely adopted. PSO is a swarm-intelligence metaheuristic algorithm inspired by collective cooperative behavior in social organisms, such as birds, fish and insects. In PSO, a swarm of particles is initialized as a set of candidate solutions to an objective function. At every iteration, each particle's current position is updated according to its previous best position, as well as the global best position found by the swarm. Due to its relatively simple implementation and gradient-free nature, PSO has been applied to a number of optimization problems ranging from image processing to electrical power systems optimization. However, it is known to be susceptible to premature convergence due to insufficient population diversity, particularly when solving highly complex and multi-modal optimization problems, as well as suboptimal particle velocity tuning which can lead to stagnation around an optimal solution. Fundamentally, PSO steers its search by a few elite positions instead of using the information held in the population as a whole. 
    
Differential Evolution (DE) \cite{storn1997differential} removes this dependence on elite attractors by deriving search directions from the population itself. DE begins by randomly initializing a set of candidate solutions represented as real-valued vectors. At each iteration, new candidate solutions, referred to as trial vectors, are generated through mutation strategies that typically involve the weighted difference of randomly selected population members. The resulting donor vectors undergo recombination with existing solutions, creating trial vectors that are evaluated according to the corresponding objective function. Solutions are then selected based on their fitness, guiding the population toward the global optimum. Despite its robustness and applicability to high-dimensional optimization problems, DE remains sensitive to noisy objective functions as well as parameter tuning, particularly regarding the mutation factor, crossover rate and population size, leading to issues such as premature convergence or inefficient search space exploration.

The Covariance Matrix Adaptation Evolution Strategy (CMA-ES) \cite{hansen2001completely} addresses this limitation by explicitly handling the dependencies between variables, which makes it one of the most robust metaheuristics for handling ill-conditioned and multimodal problems. CMA-ES optimizes a given objective function by iteratively adapting a multivariate normal mutation distribution. Initially, the mutation is modelled as a normal distribution with zero mean, characterized by a covariance matrix. At each iteration, candidate solutions are sampled from the current distribution and are evaluated according to the objective function. The mean, step size and covariance matrix of the distribution are subsequently updated according to the solutions with the highest fitness. This strategy allows the algorithm to progressively adapt its search distribution to the local geometry of the optimization landscape. Despite its strong local adaptation capabilities, CMA-ES remains vulnerable to converging to local optima on multimodal landscapes, a weakness that has been reduced by restart schemes with increasing population size, such as IPOP-CMA-ES \cite{auger2005restart}. Even so, CMA-ES shares a limitation with PSO and DE: all three operate on a fixed-length real-valued vector. They can optimise the weights or hyperparameters of a given network, but the topology must still be manually optimized.

Architecture search requires an encoding whose size can change during evolution. NeuroEvolution of Augmenting Topologies (NEAT) \cite{Kenneth2002} introduced dynamic direct genetic encodings that allow the optimization of both neural network parameters and topology. Built on principles of complexifying evolution, NEAT initializes a population of relatively simple artificial neural networks (ANNs) and progressively increases their structural complexity through mutation and crossover. Each network is encoded as a flexible genome comprising node and connection genes, enabling the emergence of new graph components while maintaining direct genotype-phenotype mapping. Historical markings are assigned to structural mutations, providing a consistent record of their evolutionary origin and allowing corresponding genes to be aligned during crossover. This framework enables architecture search and weight optimization within a dynamically expanding search space, allowing increasingly complex topologies to evolve in response to fitness-driven selection. 

Because NEAT involves direct mapping of a network’s elements to a genetic encoding, it is impractical for scaling to complex problems and modern deep architectures, such as convolutional neural networks. Indirect encodings address this limitation by evolving a compact rule that generates the network rather than the network itself. HyperNEAT \cite{stanley2009hypercube} uses NEAT to evolve a Compositional Pattern-Producing Network (CPPN) \cite{stanley2007compositional}, which computes each connection weight from the spatial coordinates of the two neurons it joins. Because a single function generates the entire connectivity, structural motifs are reused across the network allowing large architectures to emerge from small genomes. This reuse, however, comes with a bias towards regularity, and HyperNEAT’s advantage diminishes on problems that require irregular connectivity \cite{clune2011performance}. 

Existing NAS methods either optimize within predefined architectural search spaces, evolve network structures without explicitly incorporating biologically derived organizational patterns, or integrate biological knowledge as fixed architectural constraints. To address these limitations, we propose TRON, a neuroevolutionary framework that incorporates experimentally characterized PPI networks as architectural building blocks for artificial neural networks. TRON leverages biologically derived structural priors during ANN initialization, introducing connectivity patterns shaped by millions of years of biological evolution into the evolutionary search process. These architectures are subsequently optimized through mutation and fitness-based selection, allowing their connectivity to evolve in response to task-specific objectives. By combining the architectural flexibility of neuroevolution with the modular and hierarchical organization observed in biological networks, TRON investigates whether evolutionarily shaped connectivity patterns can provide a more robust substrate for neural architecture optimization. We evaluate this hypothesis by comparing TRON with established optimization and neuroevolutionary approaches across classification and regression benchmarks, assessing both predictive performance and evolutionary robustness.

\section{Material and Methods}

\subsection{Datasets} 

The primary objective of this study is not to optimize performance for a specific application domain, but rather to evaluate whether biologically derived architectural priors improve the effectiveness and robustness of neuroevolutionary search. Consequently, the experimental evaluation was designed to assess the generality of the proposed framework across multiple machine learning problems with different statistical properties. To this end, TRON was evaluated on a collection of widely used benchmark datasets comprising both classification and regression tasks.

Because TRON is intended to improve the evolutionary search process itself, rather than exploit task-specific biological information, benchmarks were intentionally selected from unrelated domains. This design minimizes the possibility that any performance improvements arise from a direct correspondence between the source biological pathways and the target prediction task. Furthermore, the selected tasks were intentionally unrelated to the biological origin of the PPI networks. Consequently, any observed improvements are more likely to reflect general properties of the evolutionary search process rather than domain-specific biological knowledge.

For classification, we selected the CORA, CITESEER, and PUBMED datasets. These citation-network benchmarks have been extensively used for the evaluation of machine learning, graph mining, and graph neural network methods, making them among the most widely adopted benchmark datasets for node classification tasks \cite{sen2008collective, yang2016revisiting, kipf2017semi}. CORA contains scientific publications categorized into seven research topics, CITESEER comprises six classes and exhibits greater sparsity and class overlap, whereas PUBMED represents a larger biomedical citation network consisting of three classes. Together, these datasets provide classification problems of increasing scale and complexity and enable the evaluation of evolved neural architectures under different topological and statistical conditions.

For regression, we selected the DIABETES, FRIEDMAN, and CALIFORNIA datasets. These benchmarks are commonly employed for evaluating regression algorithms due to their diversity in sample size, dimensionality, noise characteristics, and underlying functional relationships \cite{efron2004least, friedman1991multivariate, pace1997sparse}. The DIABETES dataset represents a real-world biomedical prediction problem involving physiological measurements, whereas the FRIEDMAN dataset is a synthetic benchmark specifically designed to assess the ability of machine learning models to capture nonlinear interactions in the presence of irrelevant variables. The CALIFORNIA housing dataset constitutes a larger real-world regression problem characterized by heterogeneous demographic and geographic predictors. Collectively, these benchmarks enable the assessment of evolved architectures across a broad range of continuous prediction scenarios.

The selected benchmarks collectively provide a broad evaluation framework spanning graph-base node classification, tabular classification, synthetic nonlinear regression, and real-world regression problems. This diversity is particularly important in the context of neuroevolution because the central hypothesis being tested concerns the evolvability of the architectural substrate itself rather than task-specific inductive biases. If protein-derived network topologies indeed provide a more favorable substrate for evolutionary optimization, performance improvements should emerge across multiple problem domains rather than being restricted to a single dataset type.

Furthermore, the selected datasets vary substantially in terms of dimensionality, complexity, and underlying data-generating processes. This heterogeneity makes them suitable for assessing not only predictive performance but also the robustness, stability, and generality of the evolutionary search process.

\subsection{TRON Framework}

\subsubsection{Protein Graph Construction}

The proposed framework constructs biologically informed graph structures from protein--protein interaction (PPI) datasets obtained from biological organisms. Each dataset is represented as an undirected interaction graph, in which proteins are represented as vertices and experimentally observed interactions are represented as edges. Formally,

$$
G=(V,E),
$$

where

\begin{itemize}
\item $V={p_1,p_2,\ldots,p_n}$ denotes the set of proteins;
\item $E\subseteq V\times V$ denotes the set of interaction edges.
\end{itemize}

The degree of each protein is defined as

$$
d(p_i)=|N(p_i)|,
$$

where $N(p_i)$ denotes the neighborhood of protein $p_i$.

Following graph construction, the interaction network is represented using an adjacency matrix that captures pairwise connectivity between proteins. Let

$$
A\in\{0,1\}^{|V|\times|V|}
$$

denote the adjacency matrix, where each entry is defined as

$$
A_{ij}=
    \begin{cases}
        1, & (p_i,p_j)\in E,\\
        0, & \text{otherwise}.
    \end{cases}
$$

To guide biologically localized topology mutations, the framework constructs hierarchical neighborhood layers through recursive graph expansion. Given an initial protein $p$, the zeroth layer is defined as

$$
L_0(p)=\{p\}.
$$

Subsequent layers are generated recursively as

$$
L_t(p)=
\left(\bigcup_{u\in L_{t-1}(p)}N(u)\right)
\setminus
\bigcup_{i=0}^{t-1}L_i(p),
$$

where $N(u)$ denotes the neighborhood of protein $u$. This construction ensures that each protein is assigned to the layer corresponding to its shortest graph distance from the initial protein.

To obtain global mutation regions, neighborhood layers are aggregated across all proteins:

$$
P_t=\bigcup_{p\in V}L_t(p).
$$

These layer pools define biologically meaningful candidate regions for topology mutation. To maintain consistency between biological locality and neural architecture depth, separate layer pools are precomputed for different network depths. Consequently, deeper neural architectures can utilize correspondingly deeper graph neighborhoods during initialization and subsequent evolutionary mutations.

\subsubsection{Graph Preprocessing}

Prior to neural topology initialization, the protein interaction graphs are preprocessed to improve structural consistency, computational efficiency, and evolutionary stability. Given an initial graph $G^{(0)}=(V^{(0)},E^{(0)})$, iterative $k$-core filtering removes proteins with insufficient interaction connectivity. At iteration $t$, the node and edge sets are updated as

\begin{equation}
V^{(t+1)}={p_i\in V^{(t)} \mid d_{G^{(t)}}(p_i)\geq k},
\end{equation}

and

\begin{equation}
E^{(t+1)}={(p_i,p_j)\in E^{(t)}\mid p_i,p_j\in V^{(t+1)}}.
\end{equation}

Here, $d_{G^{(t)}}(p_i)$ denotes the degree of protein $p_i$ in graph $G^{(t)}$. Filtering continues until convergence, i.e.,

$$
V^{(t+1)}=V^{(t)},
$$

yielding the filtered graph

$$
G'=(V',E'),
$$

which is subsequently used for neural architecture construction. This procedure removes weakly connected regions, produces denser interaction neighborhoods, and reduces the computational cost associated with topology exploration during evolution.

Following filtering, layer pools corresponding to different architecture depths are generated through iterative neighborhood expansion and cached for reuse during mutation. This avoids repeated graph traversal while preserving biologically localized topology modifications.

\subsubsection{Protein-Guided Neural Initialization}

Following graph preprocessing, the framework initializes neural architectures directly from the filtered protein interaction graph. Unlike conventional approaches that employ dense or randomly generated connectivity, the proposed method derives sparse neural topologies from biologically observed protein interactions. The initialization procedure consists of three stages:

\begin{enumerate}
\item protein-to-neuron assignment
\item connectivity initialization
\item sparse weight initialization
\end{enumerate}

Let the neural architecture be defined as

\begin{equation}
A={L_0,L_1,\ldots,L_D},
\label{eq:architecture}
\end{equation}

where $D$ denotes the network depth and $L_i$ denotes the set of neurons in layer $i$.

\begin{figure}
    \centering
    \includegraphics[width=0.99\linewidth]{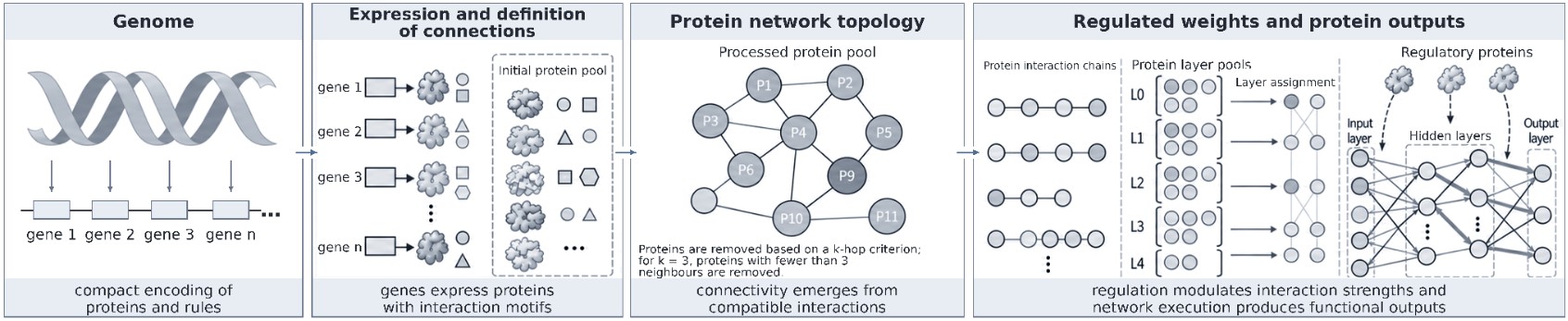}
    \caption{Protein Processing \& Network Initialization}
    \label{fig:init}
\end{figure}

For each architecture depth, the framework selects the corresponding filtered interaction graph $G_D=(V_D,E_D)$ and its precomputed layer pools. These structures define both the initial neural topology and the biologically constrained search space used during evolutionary optimization. The aforementioned process is illustrated in Figure ~\ref{fig:init}

\subsubsection{Protein-to-Neuron Mapping}

Each neuron is assigned a protein sampled from the corresponding biological layer pool. Let

$$
P_i\subseteq V_D
$$

denote the set of proteins available for assignment to neural layer $i$. The mapping is defined as

\begin{equation}
\phi:n_j^{(i)}\mapsto p_k,
\end{equation}

where $n_j^{(i)}$ denotes neuron $j$ in neural layer $i$, and $p_k$ denotes its assigned protein.

To encourage structurally informative assignments, protein sampling is weighted according to the neighborhood overlap with the subsequent layer pool:

\begin{equation}
s(p)=\left(|N(p)\cap P_{i+1}|+\epsilon\right)^3,
\end{equation}

where $N(p)$ denotes the neighbors of protein $p$, $P_{i+1}$ denotes the subsequent protein layer pool, and $\epsilon>0$ prevents zero-probability sampling. The cubic weighting biases the sampling process toward proteins with a larger number of interactions with the subsequent layer, thereby increasing their probability of being selected for network construction.

For the output layer, proteins are selected from those directly connected to proteins assigned to the preceding hidden layer. This constraint preserves biological reachability across the neural architecture.

\subsubsection{Connectivity Initialization}

Following protein assignment, sparse inter-layer connectivity is constructed directly from the biological interaction graph. For two consecutive neural layers, let

\begin{equation}
P_i={p_1^{(i)},p_2^{(i)},\ldots,p_{n_i}^{(i)}},
\qquad
P_{i+1}={p_1^{(i+1)},p_2^{(i+1)},\ldots,p_{n_{i+1}}^{(i+1)}}.
\end{equation}

The connectivity matrix is defined as

$$
C^{(i)}\in\{0,1\}^{n_i\times n_{i+1}},
$$

with entries

\begin{equation}
C_{uv}^{(i)}=
\begin{cases}
1, & (p_u^{(i)},p_v^{(i+1)})\in E_D,\\ 0, & \text{otherwise}.
\end{cases}
\end{equation}

Thus, two neurons are connected only when their assigned proteins interact within the biological network. To identify isolated neurons, the outgoing and incoming degrees are computed as

\begin{equation}
d_{\mathrm{out}}(u)=\sum_v C_{uv}^{(i)},
\qquad
d_{\mathrm{in}}(v)=\sum_u C_{uv}^{(i)}.
\end{equation}

If a neuron has no incoming or outgoing connections, a corrective edge is introduced by selecting a biologically valid interaction from the corresponding protein neighborhood. This ensures that every neuron can participate in signal propagation while maintaining the biological connectivity constraints imposed by the interaction graph.

\subsubsection{Weight Initialization}

Weights are initialized only along biologically valid neural connections. For each connectivity matrix

$$
W^{(i)}\in\mathbb{R}^{n_i\times n_{i+1}},
$$

the corresponding weights are initialized as

\begin{equation}
W_{uv}^{(i)}=
\begin{cases}
U(-\alpha,\alpha), & C_{uv}^{(i)}=1,\\
0, & C_{uv}^{(i)}=0,
\end{cases}
\end{equation}

where $U(-\alpha,\alpha)$ denotes a uniform distribution over the interval $[-\alpha,\alpha]$. The initialization scale is defined as

\begin{equation}
\alpha=
\min\left(
\frac{1}{\sqrt{n_i}},
0.1
\right).
\end{equation}

\subsection{Evolutionary Framework}

\begin{figure}[!h]
    \centering
    \includegraphics[width=0.99\linewidth]{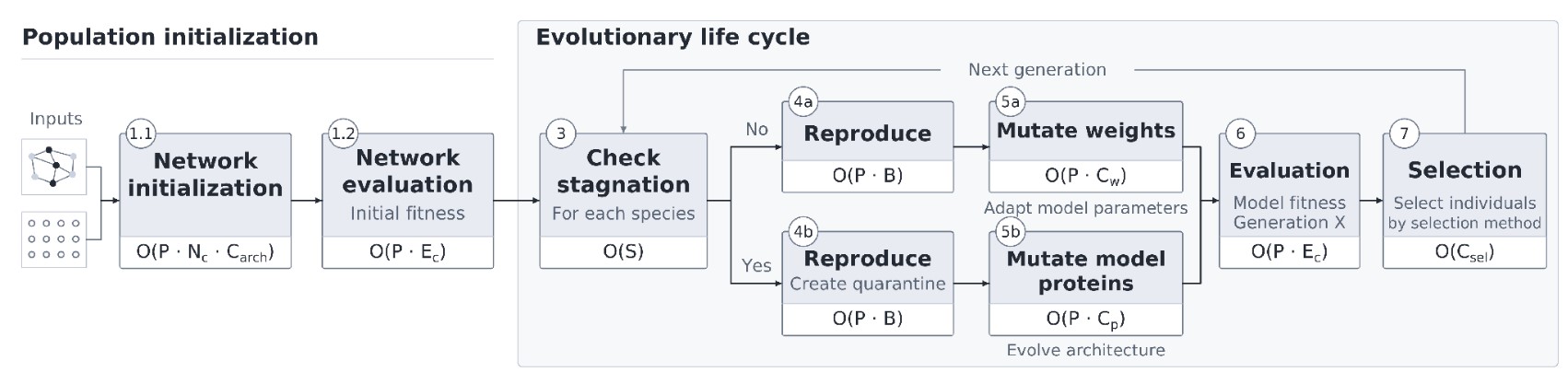}
    \caption{TRON Evolutionary Lifecycle}
    \label{fig:life}
\end{figure}

The proposed framework employs a hierarchical evolutionary strategy in which candidate neural architectures are organized according to their biological origin, structural characteristics, and evolutionary lineage. The evolutionary process combines species-level isolation, lineage-aware selection, parameter-level weight mutations, and protein-guided topology mutations. This organization allows the framework to explore multiple biologically constrained neural architectures concurrently while limiting premature convergence toward a single topology. The evolutionary lifecycle, specific stages, and computational cost per evolutionary stage are illustrated in Figure ~\ref{fig:life}.

\subsubsection{Species Definition}
The framework employs hierarchical speciation to partition the evolutionary search space according to biological origin and neural architecture. Rather than evolving a homogeneous population, individuals are organized into biologically and structurally distinct species, enabling concurrent exploration of multiple topology configurations while preserving evolutionary diversity.

\paragraph{Biological Species}

At the highest level, species are defined according to the biological source of the protein--protein interaction (PPI) network. Let $\mathcal{S}_B$ denote the set of biological species such that:

\begin{equation}
\mathcal{S}_B={S_1,S_2,\ldots,S_m}
\end{equation}

where each biological species $S_k$ corresponds to an organism-specific interaction graph

\begin{equation}
G_k=(V_k,E_k),
\end{equation}

with $V_k$ and $E_k$ denoting the corresponding proteins and interactions, respectively. Neural architectures initialized from $S_k$ inherit their protein assignments and connectivity constraints from $G_k$, thereby defining distinct biologically constrained topology search spaces.

\paragraph{Topology Species}

Within each biological species, individuals are further partitioned into topology species according to their architectural characteristics. Let 

\begin{equation}
    {T_1^{(k)},T_2^{(k)},\ldots,T_{n_k}^{(k)}}
\end{equation}

denote the topology species associated with biological species $S_k$. Individuals are grouped according to properties such as network depth, layer dimensionality, and connectivity organization. This secondary level of speciation allows structurally distinct neural architectures to be explored concurrently while reducing direct competition between substantially different topology configurations.

\paragraph{Evolutionary Population and Lineage Tracking}

Each species maintains an evolutionary population, 

\begin{equation}
P_k={m_1,m_2,\ldots,m_r},
\end{equation}

where $m_i$ denotes an individual neural model belonging to species $S_k$. Each individual is assigned a lineage identifier based on its ancestral origin, allowing evolutionary histories to be tracked throughout optimization.

For a lineage root $l_i$, its population size is defined as

\begin{equation}
\left| {m_j\in P_k\mid \operatorname{root}(m_j)=l_i}
\right|.
\end{equation}

These lineage statistics are incorporated into survivor selection to limit the dominance of individual evolutionary lineages and preserve population diversity.

\paragraph{Species Adaptation Through Stagnation Detection}

Each species independently monitors optimization progress using its best fitness, denoted by $f_k^{\mathrm{best}}(t)$. Species $S_k$ is considered stagnant when the number of generations since its last fitness improvement reaches a predefined threshold:

\begin{equation}
t-t_k^{\mathrm{improve}}\geq\tau_{\mathrm{stag}},
\end{equation}

where $t_k^{\mathrm{improve}}$ denotes the most recent generation in which the best fitness of species $S_k$ improved, and $\tau_{\mathrm{stag}}$ denotes the stagnation threshold.

When stagnation is detected, the species can trigger topology-level adaptation through protein-guided architectural mutations. This mechanism increases exploration of previously unexplored regions of the biologically constrained topology search space.

\paragraph{Quarantine Species}

To isolate individuals who have undergone topological changes through mutation from established populations, the framework employs temporary quarantine populations. Let

\begin{equation}
Q_k={q_1,q_2,\ldots,q_n}
\end{equation}

denote the quarantine population associated with species $S_k$. Quarantine individuals are generated from high-performing members of the parent species and evolve independently for a predefined cooldown period,

\begin{equation} 
T_Q = \texttt{quarantine\_cooldown}. 
\end{equation}

During this period, quarantine individuals are evaluated without immediately replacing members of the parent species. Following the cooldown period $T_Q$, a quarantine individual may be reintegrated if its fitness exceeds that of an eligible member of the parent species:

\begin{equation}
f(q_i)>f(p_j),
\end{equation}

where $q_i\in Q_k$ is a quarantine individual and $p_j\in P_k$ is an individual from the corresponding parent species.

The quarantine mechanism therefore permits large topological changes to be evaluated independently before they are allowed to influence the established evolutionary population.

\subsubsection{Reproduction}

During reproduction, a subset of each species' members is selected to generate offspring for the subsequent generation. Two primary parameters control this process: the number of individuals eligible for reproduction and the fertility assigned to each selected individual.

\begin{itemize}
    \item \textbf{Member selection:} determines the number of high-performing individuals from the current population that are eligible to reproduce. This may be implemented by selecting the top-$k$ individuals according to fitness.

    \item \textbf{Fitness:} determines the number of offspring generated by each selected individual. Each offspring initially inherits the parent's network parameters, connectivity, and protein assignments before undergoing mutation.

\end{itemize}

The configuration of these parameters determines the balance between exploitation and population diversity. Increasing the number of reproductive individuals increases the diversity of parental genome pool, whereas limiting fertility prevents excessive proliferation of closely related individuals and unnecessary growth in computational and memory requirements.

\subsubsection{Mutations}

The proposed framework optimizes biologically constrained neural architectures using gradient-free evolutionary search. Rather than relying on gradient-based optimization, candidate solutions evolve through repeated cycles of evaluation, selection, reproduction, and mutation. Individuals are organized into species and lineages, allowing structurally distinct architectures to evolve concurrently while maintaining population diversity.

The framework employs two complementary classes of mutation operators, summarized in Table~\ref{tab:mutation_summary}.

\begin{table}[h] \centering \caption{Mutation operators employed by the proposed framework.} \label{tab:mutation_summary} \begin{tabular}{ll} \hline \textbf{Mutation Type} & \textbf{Purpose} \\ \hline Weight mutation & Perturbs existing connection weights. \\ Protein mutation & Modifies protein assignments and connectivity. \\ \hline \end{tabular} \end{table}

Weight mutations provide local parameter-level exploration, whereas protein mutations modify the underlying neural topology through changes in protein assignments. Mutation intensity is adaptively adjusted according to individual fitness and species-level stagnation, providing a mechanism for balancing local parameter refinement with larger-scale structural exploration.

\subsubsection{Weight Mutations}

Weight mutation combines several complementary operators acting at different structural scales. The mutation rate is first adjusted according to the fitness of the individual:

\begin{equation}
\alpha(f) = \max\left(exp_{min},(1-f)^{1.25}\right),
\end{equation}

where $exp_{min}$ denotes the minimum exploration and $f$ denotes the model accuracy. This determines the effective mutation probability

\begin{equation}
\mu = \max\left(p_{\min},p_{\mathrm{base}}\alpha(f)\right).
\end{equation}

where $p_{min}$ and $p_{base}$ denote the minumum allowed mutation rate and the base mutation rate respectively. The core weight mutation applies multiplicative and additive Gaussian perturbations:

\begin{equation}
W\odot\exp(\epsilon_m)+\epsilon_a,
\end{equation}

followed by bounded normalization:

\begin{equation}
\frac{W'}{1+|W'|/r},
\end{equation}

where $\odot$ denotes element-wise multiplication, $r$ controls the effective weight range, and the perturbation terms are sampled according to

\[
\epsilon_m\sim\mathcal{N}(0,\sigma_m^2),
\qquad
\epsilon_a\sim\mathcal{N}
\left(
0,
\sigma_a^2(|W|+1)
\right).
\]

Here, $\sigma_m$ and $\sigma_a$ denote the standard deviations of the multiplicative and additive perturbations, respectively.

Higher-level mutation operators apply the same underlying perturbation mechanism at different structural scales. Their implementation is summarized in Table~\ref{tab:mutation_weight}.

\begin{table}[h]
\centering
\caption{Weight mutation operators.}
\label{tab:mutation_weight} 
\begin{tabular}{llll}
\hline
\textbf{Operator} &
\textbf{Target} &
\textbf{Probability} &
\textbf{Operation}\\
\hline
Element-wise & Individual weights & 0.60 &
$\mathcal{M}(W_{ij})$ \\

Neuron-wise & Incoming neuron weights & 0.25 &
$\mathcal{M}(\mathbf{w}_j)$ \\

Block & Weight submatrix & 0.15 &
$\mathcal{M}(W_B)$ \\

Layer-scale & Entire layer & 0.05 &
$sW$ \\

\hline
\end{tabular}
\end{table}

The mutation operators are sampled independently during each generation, such that multiple operators may be applied to the same individual. Consequently, the probabilities in Table~\ref{tab} represent independent activation probabilities rather than mutually exclusive categorical probabilities. The probability that none of the operators is activated is therefore

\begin{equation}
\prod_k(1-P_k).
\end{equation}

This formulation permits multiple mutation operators to act concurrently on different structural components of the same network.

\subsubsection{Protein Mutations}

Protein mutations perform topology-level adaptation by modifying the protein assignments that determine biologically constrained neural connectivity. Unlike weight mutations, protein mutations are activated when a species exhibits prolonged stagnation:

\begin{equation}
t-t_k^{\mathrm{improve}}
\geq
\tau_{\mathrm{stag}},
\end{equation}

where $\tau_{\mathrm{stag}}$ denotes the species-level stagnation threshold.

When this condition is satisfied, a temporary quarantine population $Q_k\subseteq S_k$ is generated from high-performing individuals of the parent species. Each selected individual undergoes protein reassignment according to

\begin{equation}
\phi^{(l)}\mapsto p_r,
\end{equation}

where $n_j^{(l)}$ denotes neuron $j$ in neural layer $l$ and $p_r$ denotes its assigned protein. A selected assignment is replaced according to

\begin{equation}
\phi'(n_j^{(l)})=p_r',
\qquad
p_r'\in P_l,
\end{equation}

where $P_l$ denotes the protein layer pool corresponding to neural layer $l$.

The number of protein assignments subjected to mutation is determined by

\begin{equation}
\max\left(
1,
\left\lfloor
N p_{\mathrm{prot}}
\right\rfloor
\right),
\end{equation}

where $N$ denotes the number of neurons and $p_{\mathrm{prot}}$ denotes the adaptive protein mutation rate.

Following protein reassignment, neural connectivity is reconstructed from the updated protein interaction graph. The resulting architecture is then validated to ensure forward connectivity, output reachability, and consistency with the biological interaction constraints.

Mutated individuals remain in the quarantine population for the predefined cooldown period before potential reintegration into the parent species. Superior offspring can subsequently replace weaker members of the parent population, allowing substantial topological modifications to be evaluated without immediately disrupting established evolutionary lineages.

\subsubsection{Selection and Optimization}

Evolution proceeds through repeated cycles of evaluation, reproduction, mutation, and survivor selection. Each species evolves independently while maintaining lineage information and diversity-aware selection.

An individual neural architecture is represented as

\begin{equation}
m_i=(\theta_i,C_i,P_i),
\end{equation}

where $\theta_i$, $C_i$, and $P_i$ denote the network weights, connectivity structure, and protein assignments, respectively. Following evaluation, each individual receives a task-specific fitness score:

\begin{equation}
f_i=\mathcal{F}(m_i).
\end{equation}

Selection operates at two temporal scales. During ordinary generations,

\begin{equation}
g\bmod\tau_{\mathrm{global}}\neq0,
\end{equation}

offspring compete within their respective parental lineages. For lineage $L_j$, the highest-fitness individual is given by

\begin{equation}
\arg\max_{m\in L_j} f(m).
\end{equation}

At periodic global selection intervals,

\begin{equation}
g\bmod\tau_{\mathrm{global}}=0,
\end{equation}

survivor selection is instead performed across the entire species population.

To reduce the potential for dominant evolutionary lineages to monopolize the population, global selection can use lineage-adjusted fitness:

\begin{equation}
f_i
\frac{1}{1+\alpha L_i},
\end{equation}

where

\begin{equation} 
L_i= \left| \left\{ m_j\in\mathcal P^{(t)}:r_j=r_i \right\} \right| 
\end{equation}

denotes the number of individuals in the current population that descend from the same lineage root $r_i$, and $\alpha$ controls the strength of the lineage penalty.

The framework supports multiple global survivor-selection strategies, summarized in Table~\ref{tab}.

\begin{table}[h]
\centering
\caption{Global survivor-selection strategies.}
\label{tab}
\begin{tabular}{lll}
\hline
\textbf{Strategy} & \textbf{Selection Rule} & \textbf{Characteristics}\\
\hline
Elitism &
Highest $\hat{f}_i$
&
Direct preservation of high-fitness individuals
\\

Tournament &
$\displaystyle
m^*=\arg\max_{m\in\mathcal{T}}\hat{f}(m)
$
&
Adjustable selection pressure through tournament size
\\

Rank-based &
Rank-dependent probability
&
Probabilistic selection based on population rank
\\
\hline
\end{tabular}
\end{table}

For tournament selection, the tournament size may remain fixed or vary throughout evolution. For example, selection pressure can increase linearly according to

\begin{equation}
k_{\min}
+
(k_{\max}-k_{\min})
\frac{t}{T},
\end{equation}

or geometrically according to

\begin{equation}
k_{\min}\left(\frac{k_{\max}}{k_{\min}}\right)^{t/T},
\end{equation}

where $T$ denotes the total number of generations. These schedules allow selection pressure to increase progressively during optimization.

Together, lineage-local replacement, lineage-aware fitness adjustment, and periodic global survivor selection provide complementary mechanisms for balancing exploitation of high-performing architectures with continued exploration of diverse topological configurations.

\section{Results and Discussion}

Since its introduction, NEAT has become one of the most influential neuroevolutionary frameworks by demonstrating that network topology and connection weights can be evolved simultaneously while preserving innovation through speciation \cite{Kenneth2002}. The central insight of NEAT is that architectural innovation must be protected from direct competition with already established solutions. TRON adopts this principle but extends it in two important directions. First, the search process begins from biologically derived topologies rather than minimal or randomly generated architectures. Second, the introduction of species quarantine provides an additional mechanism through which newly generated topological innovations can undergo a period of maturation before entering the broader competitive population.

From this perspective, TRON differs conceptually from most existing neuroevolution approaches. Traditional methods primarily seek improved exploration through modifications of evolutionary operators, fitness functions, diversity maintenance procedures, or encoding schemes. By contrast, TRON focuses on the nature of the structures being evolved. Rather than asking how evolution can more effectively search arbitrary architectural spaces, TRON asks whether the space itself can be made more evolvable by seeding it with organizational motifs that have already been optimized through natural evolution. This distinction is closely related to the concept of evolvability, defined as the capacity of a system to generate heritable phenotypic variation that can subsequently be refined through selection \cite{kirschner1998evolvability}. In biological systems, evolvability is frequently associated with properties such as modularity, robustness, reuse of functional components, and hierarchical organization. Protein interaction networks exhibit many of these characteristics, having emerged through extensive evolutionary refinement. Consequently, rather than constructing candidate neural architectures from random topologies, TRON begins from connectivity motifs that have already persisted throughout biological evolution.

To determine whether these biologically derived architectural priors provide an advantage during evolutionary search, we evaluated TRON on a diverse collection of classification and regression benchmarks and compared its performance against several established evolutionary and neuroevolutionary approaches. Because the selected tasks are unrelated to the biological origin of the protein networks, any observed improvements are more likely to reflect general properties of the evolutionary search process rather than domain-specific biological knowledge.

\begin{table}[H]
\caption{Algorithm performance on CORA}
\label{tab:cora-performance}
\centering
\setlength{\tabcolsep}{6pt}
\renewcommand{\arraystretch}{1.2}
\begin{tabular}{lcccccc}
\toprule
\multirow{2}{*}{\textbf{Method}} &
\multicolumn{6}{c}{\textbf{CORA}} \\
\cmidrule(lr){2-7}
& \textbf{MIN} & \textbf{MAX} & \textbf{MEDIAN}
& \textbf{MEAN} & \textbf{SD} & \textbf{SNR (dB)} \\
\midrule
CMA-ES & \textbf{0.6679} & \textbf{0.8118} & \textbf{0.7634} & \textbf{0.7429} & \textbf{0.0554} & \textbf{22.5484} \\
ES-MLP & \textbf{0.6015} & \textbf{0.7989} & \textbf{0.7386} & \textbf{0.7336} & \textbf{0.0433} & \textbf{24.5794} \\
DE     & \textbf{0.5119} & \textbf{0.5966} & \textbf{0.5527} & \textbf{0.5509} & \textbf{0.0264} & \textbf{26.3894} \\
PSO    & \textbf{0.4207} & \textbf{0.5572} & \textbf{0.4841} & \textbf{0.4779} & \textbf{0.0415} & \textbf{21.2258} \\
NEAT   & \textbf{0.4649} & \textbf{0.5793} & \textbf{0.5240} & \textbf{0.5189} & \textbf{0.0378} & \textbf{22.7518} \\
TRON   & \textbf{0.8044} & \textbf{0.8357} & \textbf{0.8293} & \textbf{0.8262} & \textbf{0.0097} & \textbf{38.6063} \\
\bottomrule
\end{tabular}%
\end{table}

\begin{table}[H]
\caption{Algorithm performance on CITESEER}
\label{tab:citeseer-performance}
\centering
\setlength{\tabcolsep}{6pt}
\renewcommand{\arraystretch}{1.2}
\begin{tabular}{lcccccc}
\toprule
\multirow{2}{*}{\textbf{Method}} &
\multicolumn{6}{c}{\textbf{CITESEER}} \\
\cmidrule(lr){2-7}
& \textbf{MIN} & \textbf{MAX} & \textbf{MEDIAN}
& \textbf{MEAN} & \textbf{SD} & \textbf{SNR (dB)} \\
\midrule
CMA-ES & \textbf{0.6698} & \textbf{0.7372} & \textbf{0.7057} & \textbf{0.7065} & \textbf{0.0205} & \textbf{30.7472} \\
ES-MLP & \textbf{0.5691} & \textbf{0.7282} & \textbf{0.7009} & \textbf{0.6785} & \textbf{0.0523} & \textbf{22.2610} \\
DE     & \textbf{0.5210} & \textbf{0.6377} & \textbf{0.6055} & \textbf{0.5919} & \textbf{0.0388} & \textbf{23.6683} \\
PSO    & \textbf{0.4565} & \textbf{0.5631} & \textbf{0.5248} & \textbf{0.5177} & \textbf{0.0340} & \textbf{23.6520} \\
NEAT   & \textbf{0.5961} & \textbf{0.6396} & \textbf{0.6239} & \textbf{0.6216} & \textbf{0.0140} & \textbf{32.9477} \\
TRON   & \textbf{0.7207} & \textbf{0.7357} & \textbf{0.7222} & \textbf{0.7250} & \textbf{0.0053} & \textbf{42.7212} \\
\bottomrule
\end{tabular}%
\end{table}

\begin{table}[H]
\caption{Algorithm performance on PUBMED}
\label{tab:pubmed-performance}
\centering
\setlength{\tabcolsep}{6pt}
\renewcommand{\arraystretch}{1.2}
\begin{tabular}{lcccccc}
\toprule
\multirow{2}{*}{\textbf{Method}} &
\multicolumn{6}{c}{\textbf{PUBMED}} \\
\cmidrule(lr){2-7}
& \textbf{MIN} & \textbf{MAX} & \textbf{MEDIAN}
& \textbf{MEAN} & \textbf{SD} & \textbf{SNR (dB)} \\
\midrule
CMA-ES & \textbf{0.6798} & \textbf{0.8141} & \textbf{0.7868} & \textbf{0.7728} & \textbf{0.0476} & \textbf{24.2092} \\
ES-MLP & \textbf{0.6841} & \textbf{0.8126} & \textbf{0.7318} & \textbf{0.7396} & \textbf{0.0692} & \textbf{20.5778} \\
DE     & \textbf{0.6526} & \textbf{0.7247} & \textbf{0.6747} & \textbf{0.6830} & \textbf{0.0228} & \textbf{29.5297} \\
PSO    & \textbf{0.6385} & \textbf{0.6992} & \textbf{0.6668} & \textbf{0.6667} & \textbf{0.0221} & \textbf{29.5908} \\
NEAT   & \textbf{0.6455} & \textbf{0.7784} & \textbf{0.7595} & \textbf{0.7508} & \textbf{0.0379} & \textbf{25.9377} \\
TRON   & \textbf{0.8242} & \textbf{0.8395} & \textbf{0.8315} & \textbf{0.8313} & \textbf{0.0046} & \textbf{45.1400} \\
\bottomrule
\end{tabular}%
\end{table}

\begin{table}[H]
\caption{Algorithm performance on DIABETES}
\label{tab:diabetes-performance}
\centering
\setlength{\tabcolsep}{6pt}
\renewcommand{\arraystretch}{1.2}
\begin{tabular}{lcccccc}
\toprule
\multirow{2}{*}{\textbf{Method}} &
\multicolumn{6}{c}{\textbf{DIABETES}} \\
\cmidrule(lr){2-7}
& \textbf{MIN} & \textbf{MAX} & \textbf{MEDIAN}
& \textbf{MEAN} & \textbf{SD} & \textbf{SNR (dB)} \\
\midrule
CMA-ES & \textbf{-0.2606} & \textbf{0.7915} & \textbf{0.2846} & \textbf{0.3037} & \textbf{0.3266} & \textbf{-0.6314} \\
ES-MLP & \textbf{-0.0553} & \textbf{0.8840} & \textbf{0.3336} & \textbf{0.3168} & \textbf{0.2904} & \textbf{0.7558} \\
DE     & \textbf{-0.0886} & \textbf{0.8144} & \textbf{0.4316} & \textbf{0.3854} & \textbf{0.3135} & \textbf{1.7935} \\
PSO    & \textbf{0.0989} & \textbf{0.8312} & \textbf{0.4307} & \textbf{0.4116} & \textbf{0.2361} & \textbf{4.8276} \\
NEAT   & \textbf{0.2586} & \textbf{0.7855} & \textbf{0.4538} & \textbf{0.4046} & \textbf{0.2246} & \textbf{5.1123} \\
TRON   & \textbf{0.5586} & \textbf{0.5676} & \textbf{0.5601} & \textbf{0.5614} & \textbf{0.0034} & \textbf{44.3559} \\
\bottomrule
\end{tabular}%
\end{table}

\begin{table}[H]
\caption{Algorithm performance on FRIEDMAN}
\label{tab:friedman-performance}
\centering
\setlength{\tabcolsep}{6pt}
\renewcommand{\arraystretch}{1.2}
\begin{tabular}{lcccccc}
\toprule
\multirow{2}{*}{\textbf{Method}} &
\multicolumn{6}{c}{\textbf{FRIEDMAN}} \\
\cmidrule(lr){2-7}
& \textbf{MIN} & \textbf{MAX} & \textbf{MEDIAN}
& \textbf{MEAN} & \textbf{SD} & \textbf{SNR (dB)} \\
\midrule
CMA-ES & \textbf{0.7449} & \textbf{0.8888} & \textbf{0.8189} & \textbf{0.8211} & \textbf{0.0507} & \textbf{24.1878} \\
ES-MLP & \textbf{0.7816} & \textbf{0.8965} & \textbf{0.8269} & \textbf{0.8242} & \textbf{0.0390} & \textbf{26.4994} \\
DE     & \textbf{0.7305} & \textbf{0.8731} & \textbf{0.7965} & \textbf{0.7959} & \textbf{0.0437} & \textbf{25.2075} \\
PSO    & \textbf{0.6698} & \textbf{0.8210} & \textbf{0.7411} & \textbf{0.7432} & \textbf{0.0528} & \textbf{22.9694} \\
NEAT   & \textbf{0.6328} & \textbf{0.8067} & \textbf{0.7284} & \textbf{0.7341} & \textbf{0.0502} & \textbf{23.3010} \\
TRON   & \textbf{0.8716} & \textbf{0.9205} & \textbf{0.8970} & \textbf{0.8962} & \textbf{0.0148} & \textbf{35.6429} \\
\bottomrule
\end{tabular}%
\end{table}

\begin{table}[H]
\caption{Algorithm performance on CALIFORNIA}
\label{tab:california-performance}
\centering
\setlength{\tabcolsep}{6pt}
\renewcommand{\arraystretch}{1.2}
\begin{tabular}{lcccccc}
\toprule
\multirow{2}{*}{\textbf{Method}} &
\multicolumn{6}{c}{\textbf{CALIFORNIA}} \\
\cmidrule(lr){2-7}
& \textbf{MIN} & \textbf{MAX} & \textbf{MEDIAN}
& \textbf{MEAN} & \textbf{SD} & \textbf{SNR (dB)} \\
\midrule
CMA-ES & \textbf{0.7121} & \textbf{0.7603} & \textbf{0.7291} & \textbf{0.7288} & \textbf{0.0184} & \textbf{31.9558} \\
ES-MLP & \textbf{0.6252} & \textbf{0.6936} & \textbf{0.6659} & \textbf{0.6656} & \textbf{0.0217} & \textbf{29.7351} \\
DE     & \textbf{0.5572} & \textbf{0.7437} & \textbf{0.6511} & \textbf{0.6569} & \textbf{0.0495} & \textbf{22.4579} \\
PSO    & \textbf{0.6068} & \textbf{0.7088} & \textbf{0.6643} & \textbf{0.6667} & \textbf{0.0355} & \textbf{25.4740} \\
NEAT   & \textbf{0.4899} & \textbf{0.5962} & \textbf{0.5618} & \textbf{0.5488} & \textbf{0.0400} & \textbf{22.7471} \\
TRON   & \textbf{0.7044} & \textbf{0.7527} & \textbf{0.7378} & \textbf{0.7336} & \textbf{0.0177} & \textbf{32.3497} \\
\bottomrule
\end{tabular}%
\end{table}

\subsection{Classification Performance} 

Classification experiments were conducted using the CORA, CITESEER, and PUBMED datasets. Across all three benchmarks, TRON consistently achieved the highest predictive performance among the evaluated evolutionary approaches (Tables~\ref{tab:cora-performance}, \ref{tab:citeseer-performance}, and \ref{tab:pubmed-performance}). The improvement was particularly pronounced for CORA and PUBMED, where TRON outperformed both classical evolutionary optimizers, including PSO, DE, and CMA-ES, and dedicated neuroevolutionary approaches such as NEAT. Similar advantages were observed for CITESEER despite the greater difficulty associated with the dataset. 

These results suggest that biologically derived connectivity patterns can substantially improve the efficiency with which evolution explores the neural architecture search space. Rather than constructing architectures entirely through random variation, TRON begins from topological organizations that already exhibit coherent interaction patterns. 

Although the specific mechanisms responsible for these improvements remain to be fully characterized, the results are consistent with the hypothesis that naturally evolved network organizations represent a more favorable substrate for evolutionary optimization than randomly generated topologies. 

An additional and particularly noteworthy observation concerns the consistency of the evolutionary process. Across all classification datasets, TRON exhibited substantially lower variability across independent runs than competing approaches. Standard deviations were often several-fold lower than those observed for alternative methods, indicating that evolutionary trajectories converged toward high-performing solutions more reliably and reproducibly. This reduction in variability may be as important as the observed gains in average predictive performance. Many optimization algorithms can occasionally discover highly effective solutions through favorable stochastic trajectories. In contrast, consistently converging toward high-quality solutions across independent runs is substantially more difficult. 

The results therefore suggest that the biological architectural priors employed by TRON constrain the search process toward more coherent regions of the architectural landscape, reducing sensitivity to random initialization and stochastic evolutionary events. More broadly, these observations support the idea that protein interaction networks may contribute not only useful initial topologies but also organizational properties that influence how effectively evolutionary operators act upon those topologies. Randomly generated architectures may be particularly vulnerable to the disruptive effects of mutation and recombination. In contrast, naturally evolved interaction networks may contain forms of modularity, robustness, or redundancy that allow beneficial structural features to persist across generations. This interpretation is consistent with previous work linking modularity and robustness to improved evolutionary adaptation. Modular systems tend to localize the effects of perturbations, thereby allowing innovations to accumulate without disrupting existing functionality \cite{wagner2008robustness}. Studies in evolutionary computation have similarly suggested that preserving functional building blocks can facilitate long-term adaptation by reducing the destructive effects of genetic variation. The protein interaction networks used in TRON may naturally provide such building blocks, allowing evolution to exploit organizational principles that have already demonstrated their effectiveness in living systems.

While the present study does not directly establish the mechanistic basis of these effects, the improved convergence stability observed throughout the classification benchmarks is consistent with this interpretation.

\subsection{Regression Performance} 

To determine whether the observed benefits extend beyond classification problems, TRON was further evaluated on three regression benchmarks: DIABETES, FRIEDMAN, and CALIFORNIA. 

On the DIABETES dataset, TRON achieved a substantial improvement over all competing methods (Table~\ref{tab:diabetes-performance}), exceeding the strongest baseline by approximately 36\%. This result demonstrates that the advantages associated with biologically derived architectural priors are not restricted to discrete classification tasks and can generalize effectively to continuous prediction problems. As observed for the classification benchmarks, the performance gain was accompanied by remarkably low variability across runs, reinforcing the view that TRON promotes stable evolutionary optimization. 

The framework also achieved the highest predictive performance on the synthetic FRIEDMAN benchmark (Table~\ref{tab:friedman-performance}). Because FRIEDMAN was specifically designed to evaluate the ability of machine learning models to capture nonlinear interactions in complex search spaces, this result suggests that the benefits of protein-derived architectures are not limited to biological datasets. Instead, it indicates that the organizational principles embedded in naturally evolved interaction networks may provide generally useful inductive biases for evolutionary architecture search. 

The CALIFORNIA housing benchmark produced a more nuanced outcome (Table~\ref{tab:california-performance}). Although TRON remained among the highest-performing approaches, its advantage over the strongest competitors was considerably smaller than that observed on the previous datasets. This observation provides an important counterbalance to the otherwise consistent performance gains observed throughout the study. Rather than suggesting a fundamental limitation of the framework, it indicates that the utility of biological architectural priors is likely context dependent. Certain predictive problems may align more naturally with the organizational principles embodied within protein interaction networks, whereas others may derive more modest benefits from such structures. 

Taken together, the regression experiments reveal a pattern similar to that observed for classification. Across datasets with distinct statistical characteristics, sample sizes, and underlying functional relationships, TRON either matched or exceeded the performance of all evaluated evolutionary competitors while maintaining superior robustness and reproducibility. 

\subsection{Biological Priors and Evolvability}

A central hypothesis motivating this work is that protein interaction networks may represent architectural building blocks that are inherently more evolvable than randomly generated topologies. The results obtained across both classification and regression tasks are broadly consistent with this hypothesis. 

Traditional neuroevolution methods primarily focus on improving the evolutionary process itself through modifications of mutation operators, crossover mechanisms, selection strategies, or diversity-preservation techniques. In contrast, TRON focuses on the substrate upon which these mechanisms operate. Rather than asking how evolution can more effectively search arbitrary architectural spaces, TRON asks whether the architectural space itself can be enriched with structures that have already undergone extensive evolutionary refinement. From this perspective, the superior performance and reduced variability exhibited by TRON may arise because protein interaction networks encode organizational properties that promote robustness to evolutionary perturbation. If useful functional structures are more likely to survive mutation and recombination, evolutionary adaptation can proceed more efficiently because beneficial modifications accumulate rather than being repeatedly disrupted. The markedly lower variance across independent runs is particularly consistent with this interpretation. It is important to note, however, that the present results do not fully disentangle the individual contributions of biological priors, speciation, and quarantine. Consequently, the observed improvements should not be attributed exclusively to any single component of the framework. 

Future ablation studies will be necessary to quantify the specific contribution of each mechanism and to identify the precise factors responsible for the enhanced robustness observed throughout the experiments. 

\subsection{Computational Trade-Offs} 

\begin{figure}[htbp]
    \centering
    \subfloat[Classification Tasks]{\includegraphics[width=0.48\textwidth]{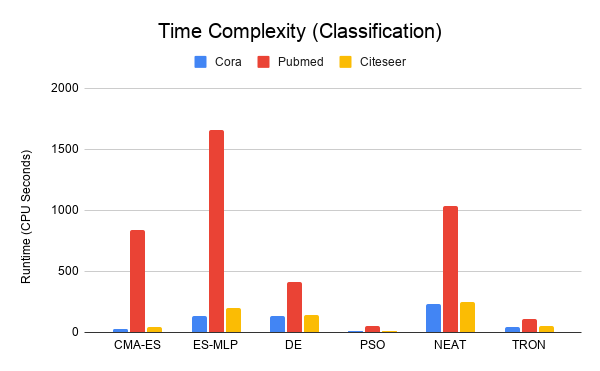}}
    \hfill
    \subfloat[Regression Tasks]{\includegraphics[width=0.48\textwidth]{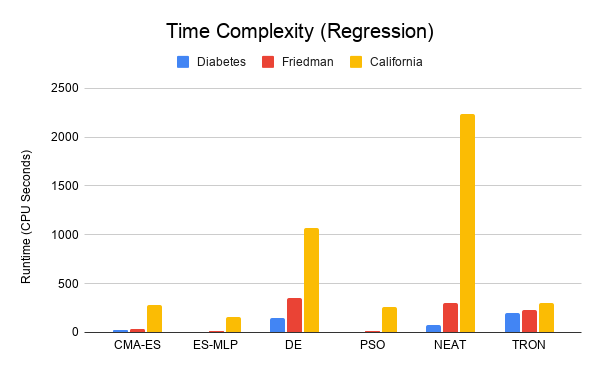}}
    \caption{Time complexity comparison for Evolutionary Frameworks quantified as CPU seconds}
    \label{fig:time_complexity}    
\end{figure}

\begin{figure}[htbp]
    \centering
    \subfloat[Classification Tasks]{\includegraphics[width=0.48\textwidth]{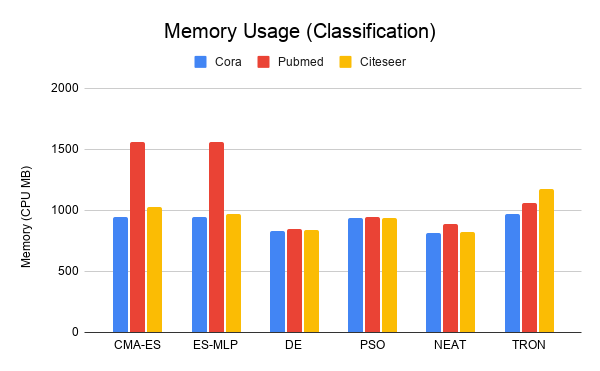}}
    \hfill
    \subfloat[Regression Tasks]{\includegraphics[width=0.48\textwidth]{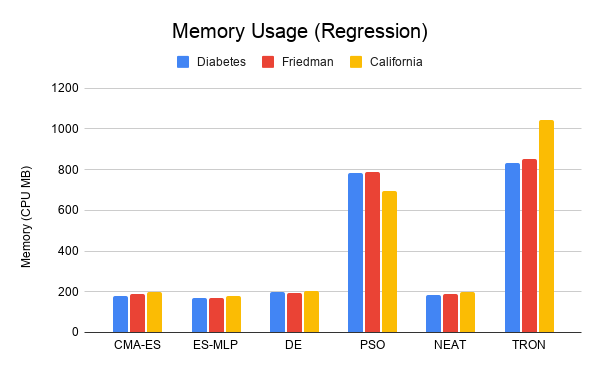}}
    \caption{Memory Usage comparison for Evolutionary Frameworks}
    \label{fig:memory_usage}
\end{figure}

The improvements in predictive performance and evolutionary stability were associated with increased computational requirements. As shown in Figures~\ref{fig:time_complexity} and~\ref{fig:memory_usage}, TRON generally requires greater memory consumption than most of the competing approaches. This increase is expected because biologically derived interaction networks have larger memory requirements in general since they retain information about the proteins that have been assigned to the respective neurons. Furthermore, the quarantine mechanism introduces additional evaluations that increase the computational burden of the evolutionary process. On the other hand, regarding time complexity, TRON is amongst the top performers, especially when tested on large datasets like Pubmed and California. At the same time, TRON requires significantly fewer runs to verify its results, courtesy of its high SNR. From an engineering perspective, this trade-off reflects a deliberate design choice. The primary goal of TRON is to improve evolvability and search robustness rather than minimize computational cost. Consequently, the framework sacrifices some degree of efficiency in exchange for a richer architectural substrate and a more stable evolutionary search process. 

Future work should investigate pathway compression strategies, sparse architectural representations, and distributed evolutionary approaches to improve scalability while retaining the benefits of biologically derived priors. 

\subsection{Limitations and Future Directions} 

Several limitations should be acknowledged. First, the present study focused exclusively on protein interaction networks and did not evaluate alternative biological network classes such as gene regulatory networks, signaling pathways, metabolic networks, or developmental programs. Second, the current experiments do not fully disentangle the contributions of biological priors, speciation, and quarantine. Third, although the results are consistent with the hypothesis that biological architectures improve evolvability, the specific structural properties responsible for this effect remain unknown. 

Despite these limitations, the findings provide encouraging evidence that naturally evolved biological networks can serve as useful architectural priors for neuroevolution. In particular, the observed improvements in robustness and reproducibility suggest that biological organization may offer advantages extending beyond predictive performance alone.

Finally, the current implementation of TRON relies exclusively on experimentally characterized protein interaction networks. Recent advances in protein structure and interaction prediction, particularly through AlphaFold and related approaches, suggest the possibility of moving beyond known interaction maps toward predicted interaction landscapes derived directly from sequence information. Future generations of TRON may therefore exploit not only experimentally validated protein networks but also vast numbers of biologically plausible interactions inferred computationally. Under this perspective, genomes themselves become repositories of potential computational architectures, enabling neuroevolution to explore a substantially larger design space derived from the products of natural evolution. 

More broadly, these findings support the possibility that real biological networks may serve as reusable computational substrates for artificial intelligence and provide a foundation for progressively transferring increasingly complex forms of biological organization into artificial systems.

\subsection*{Concluding Remarks} 

The present implementation of TRON relies on experimentally derived protein interaction networks to define the architectural substrate upon which neuroevolution operates. While this strategy already enables the direct reuse of connectivity patterns shaped through natural selection, it likely represents only a small fraction of the architectural information accessible from biological systems. 

Recent advances in protein structure prediction have dramatically expanded our ability to infer molecular interactions directly from sequence information. In particular, AlphaFold and its multimer extensions have demonstrated the ability not only to predict protein structures with near-experimental accuracy, but also to identify and model protein--protein interactions from sequence alone \cite{jumper2021highly, bryant2022improved}. Rather than restricting TRON to currently known protein interaction networks, future implementations may therefore generate architectural priors from predicted interaction landscapes derived directly from genomic sequences.

Such an approach would fundamentally expand the search space available to neuroevolution. Instead of reusing only experimentally characterized pathways, artificial neural architectures could be generated from millions of potential molecular interactions spanning diverse organisms, evolutionary lineages, and even synthetic sequences. In this scenario, evolution would no longer be limited to selecting among known biological networks, but could instead explore a vastly richer space of biologically plausible organizational motifs inferred computationally. 

More speculatively, advances in protein interaction prediction may ultimately allow the direct translation of genomic information into adaptive computational architectures. Under this perspective, genomes become encodings of network organization from which artificial systems can derive architectural priors without requiring prior biological annotation. As protein structure and interaction prediction continue to improve, the boundary between biological sequence information and computational architecture design may progressively diminish, opening new opportunities for transferring the products of natural evolution into artificial intelligence systems.


\bibliography{sample}

\end{document}